# Closing the loop on multisensory interactions: A neural architecture for multisensory causal inference and recalibration

Jonathan Tong[1], German I. Parisi[2], Stefan Wermter[2] and Brigitte Röder[1]

1.  Biological Psychology and Neuropsychology, University of Hamburg, 20146 Hamburg, Germany

2.  Department of Informatics, University of Hamburg, 22527 Hamburg, Germany

Author notes:
This research was supported by funding from the German Research Foundation (TRR 169). No conflicts of interest, financial or otherwise, are declared by the authors.

The code to implement the network model and simulations can be found online at:
https://github.com/jonathan-tong/multisensory-network-model

# Closing the loop on multisensory interactions: A neural architecture for multisensory causal inference and recalibration

**Abstract**

When the brain receives input from multiple sensory systems, it is faced with the question of whether it is appropriate to process the inputs in combination, as if they originated from the same event, or separately, as if they originated from distinct events. Furthermore, it must also have a mechanism through which it can keep sensory inputs calibrated to maintain the accuracy of its internal representations. We have developed a neural network architecture capable of i) approximating optimal multisensory spatial integration, based on Bayesian causal inference, and ii) recalibrating the spatial encoding of sensory systems. The architecture is based on features of the dorsal processing hierarchy, including the spatial tuning properties of unisensory neurons and the convergence of different sensory inputs onto multisensory neurons. Furthermore, we propose that these unisensory and multisensory neurons play dual roles in i) encoding spatial location as separate or integrated estimates and ii) accumulating evidence for the independence or relatedness of multisensory stimuli. We further propose that top-down feedback connections spanning the dorsal pathway play key a role in recalibrating spatial encoding at the level of early unisensory cortices. Our proposed architecture provides possible explanations for a number of human electrophysiological and neuroimaging results and generates testable predictions linking neurophysiology with behaviour.

# 1. Introduction

An important task constantly being carried out by the brain is determining whether inputs from different sensory modalities originate from the same cause or from separate causes. Take, for example, a situation in which multiple people in a room are having a conversation. To follow the conversation, one must identify and locate who is talking at any given moment. Considering only the spatiotemporal properties of current sensory information, the speaker whose apparent image is nearest the apparent source of the sound is the most likely candidate for who is speaking. However, taking further contextual information into consideration could enhance one's estimate for where, and thus whom, the speech is coming from: for example, the sight of moving lips combined with speech sounds further disambiguates and greatly influences the final estimate or percept of where the sound is occurring. This prior or contextual information, although helpful in most situations, can sometimes lead to mistaken perceptions, such as when a ventriloquist convinces her audience that a puppet is speaking by moving its mouth in synchrony with her own vocalizations. Because of this famous example, the general phenomenon of visual dominance over auditory cues for localization has been labeled the ventriloquism effect.

## 1.1 A review of Causal Inference

In general, the problem outlined above is a question of hierarchical causal inference (Shams & Beierholm, 2010): lower-level features, such as stimulus locations and onset times, inform the higher-level hypotheses about the causal relations between stimuli (are they related or not?). In turn, the inferences made about the causal relations affect the final estimates of where and when the stimuli occurred. Körding et al. (2007) formally introduced causal inference as a mechanism for optimal multisensory interactions. Rather than naively integrating sensory cues, under the assumption that they are related (Ernst and Banks, 2002; Alais and Burr, 2004; Atkins et al., 2001), their model assigns uncertainty to the exhaustive and mutually exclusive possibilities that i) the stimuli are causally linked by a common source (C=1), and drawn from distributions centered at the same true location or ii) that they have separate causes (C=2), and are independently drawn from their distributions centered at their own true locations (Körding et al., 2007). The appropriate strategy when dealing with these two possibilities is to first calculate the likelihood of obtaining the auditory and visual sensory estimates $(x_A, x_V)$ given either possible causal model: $p(x_v, x_a | C = 1)$ and $p(x_v, x_a | C = 2)$. The posterior probabilities of each causal model, $p(C = 1 | x_v, x_a)$ and $p(C = 2 | x_v, x_a)$, are then determined by entering the likelihoods for each causal model and the prior probability that two stimuli are causally related, $p(C = 1)$ or $p_{common}$ , into Bayes' Equation. Finally, an observer should use these posteriors as weights to compute a weighted average between the fully integrated estimate for stimulus location (assumption of a common cause) and the estimate produced under the assumption of independent stimulus locations.

A number of behavioural studies suggest that the brain appears to do causal inference and produces estimates that are well fit by the model (Körding et al., 2007; Sato, Toyoizumi & Aihara, 2007; Wozny & Shams, 2010; Odegaard, Wozny & Shams, 2015; McGovern et al., 2016), but how exactly would neurons carry out these computations? Early models have been proposed to describe how idealized neurons might carry out such computations in a Bayes optimal manner (Ma and Rahmati, 2013; Spratling, 2016; Magosso, Cuppini, and Ursino, 2017). Such models tend to focus on achieving optimality through encoding the full probability distributions over the variable of interest and less so the actual physiological or anatomical relationships observed between neural structures. Taking inspiration from such models, especially the architecture of Spratling (2016), as well as what is known from the literature regarding the neural correlates of multisensory integration, we propose a network model that approximates the average estimates produced by the causal inference model. Furthermore, our model accounts for the recalibration of sensory estimates, and considers the possible functional roles for different brain regions and their associated activities involved in multisensory causal inference and recalibration. In the next section we will review some of the major findings in the domain of the neural architecture of multisensory integration and recalibration that have informed this work.

## 1.2 Cortical structures and correlates of multisensory causal inference and recalibration

Since Meredith and Stein's (1986) pioneering work exploring single neuron responses to multisensory stimuli, a number of studies have elucidated the possible involvement of cortical areas in auditory and visual integration. Early sensory cortices, such as auditory and visual cortex, which were previously believed to have been strictly unisensory have been found to have multisensory responses (McDonald et al., 2013; Feng et al., 2014; Bieler et al., 2017; Brang et al., 2015). Although there exists direct anatomical connections between the auditory and the visual cortices (Beer et al., 2011), there is convincing evidence that much of the cross-modality effects are mediated by top-down projections from higher multisensory areas in the processing hierarchy (e.g. superior temporal sulcus (STS), or intra-parietal (IP) cortex and not necessarily from direct connections (Schröder & Foxe, 2002). In the current study, we propose a role for both feedforward convergence of unisensory projections to multisensory units and for feedback projections to lower unisensory areas.

A number of human electrophysiological and neuroimaging studies have been carried out during multisensory localization tasks. Bonath et al. (2007), and later Bruns and Röder (2010), identified a correlation between the occurrence of the ventriloquism effect and the amplitude of an event-related potential (ERP) at 260ms post stimulus (N260). When a visual stimulus biases the localization of an auditory stimulus, there is a larger N260 amplitude on the hemisphere contralateral to the side where the sound was perceived. This finding was corroborated by an fMRI analysis with the same behavioural task, which showed that the Planum Temporale (PT, an area in auditory cortex) showed higher activation on the side contralateral to the perceived sound (Bonath et al., 2007). These findings also seem to align with what is known about the dominance of contralaterally tuned auditory neurons in the auditory cortices (Werner-Reiss & Groh, 2008).

Similar studies that combine localization behavior and electrophysiology or neuroimaging have also identified a number of neural correlates of multisensory recalibration. Audio-visual recalibration is often studied as the manifestation of the ventriloquism aftereffect: after repeated exposure to an audiovisual (AV) spatial disparity (adaptation), the localization of subsequent unisensory auditory stimuli is biased in the direction of the previously presented discrepant visual stimulus. Recent studies have shown that this ventriloquism aftereffect is cumulative but can also occur after a single presentation of AV disparity (Bruns & Röder, 2015; Bosen et al., 2016), and the effect fully decays tens of seconds following the last AV stimulation (Bosen et al., 2016). The effect seems to generalize across space for at least as far as 15 degrees from the original adaptation stimulus (Bosen et al., 2016), and the effect becomes specific to sound frequency after long-term exposure to disparities (Recazone, 1998; Bruns & Röder, 2015). The latter finding has prompted both groups of authors to hypothesize that the ventriloquism aftereffect has its origins in the auditory cortices where sound processing of frequencies remain topographically organized, a theory that is supported by recordings from A1 of macaque monkeys (Recazone, 1998). A handful of other studies have lent further support to this idea: an early ERP between 70 and 130 ms post-stimulus (N1), linked to auditory cortical activity, was shown to be modulated by the occurrence of the ventriloquism aftereffect. This finding is quite surprising considering that the aftereffect and on-line ventriloquism effect have different ERP correlates, and that the aftereffect is linked to an earlier component (N1) than the on-line effect (N260). It has been suggested that early ERP components are linked to feedforward activation, while later ERP components are linked to top-down feedback activation (Garrido et al. 2007).

We propose a network model that incorporates the findings we have just reviewed. The model contains a neural architecture that captures the convergence of unisensory auditory and visual input in multisensory areas, such as the IP. As a population, these multisensory units represent how probable it is that the input came from the same external cause, as well as the appropriate spatial estimate given this assumption (a fully integrated estimate). The model also accounts for unisensory auditory and visual units found in these multisensory areas, assigning them an essential role in causal inference: they represent the location estimates for each individual sensory system under the assumption that they are independent. The model further illustrates how auditory and visual responses with very different spatial representations (rate versus place code) could be simply combined to infer spatial location. Lastly, by considering the feedback connections between IP and unisensory cortices, we implemented a mechanism for audiovisual recalibration that accounts for the ventriloquism aftereffect and its time course of accumulation and decay. In the next section we will describe, in formal terms, the neural architecture for multisensory causal inference and unsupervised recalibration designed around our knowledge of the reviewed literature.

## 2. Methods: A Description of the Model

The model is a 3-layer network consisting of: i) an input layer of unisensory units with response properties modeled after neurophysiological recordings from primate auditory and visual cortical neurons (Werner-Reiss & Groh, 2008), ii) an intermediate layer (functionally resembling IP) containing a subset of multisensory units which receive inputs from both auditory and visual units, and a subset of unisensory units which receive inputs from only one sensory modality, and iii) an output or reconstruction layer containing subsets of units that reconstructs the auditory and visual inputs. The model makes use of top-down feedback connections from the output/reconstruction layer to the input layer in order to readjust input weights in an adaptive manner. For an illustration of the network architecture, see *figure 1*. In all simulations reported here, we used a fixed number of units in each population: $n = 301$ units, spanning a range of preferred/central locations of -150 to 150 degrees in steps of 1 degree. We covered a range well beyond 90 degrees in order to minimize the frequency of boundary effects. For a list of all parameter settings see *table 1*.

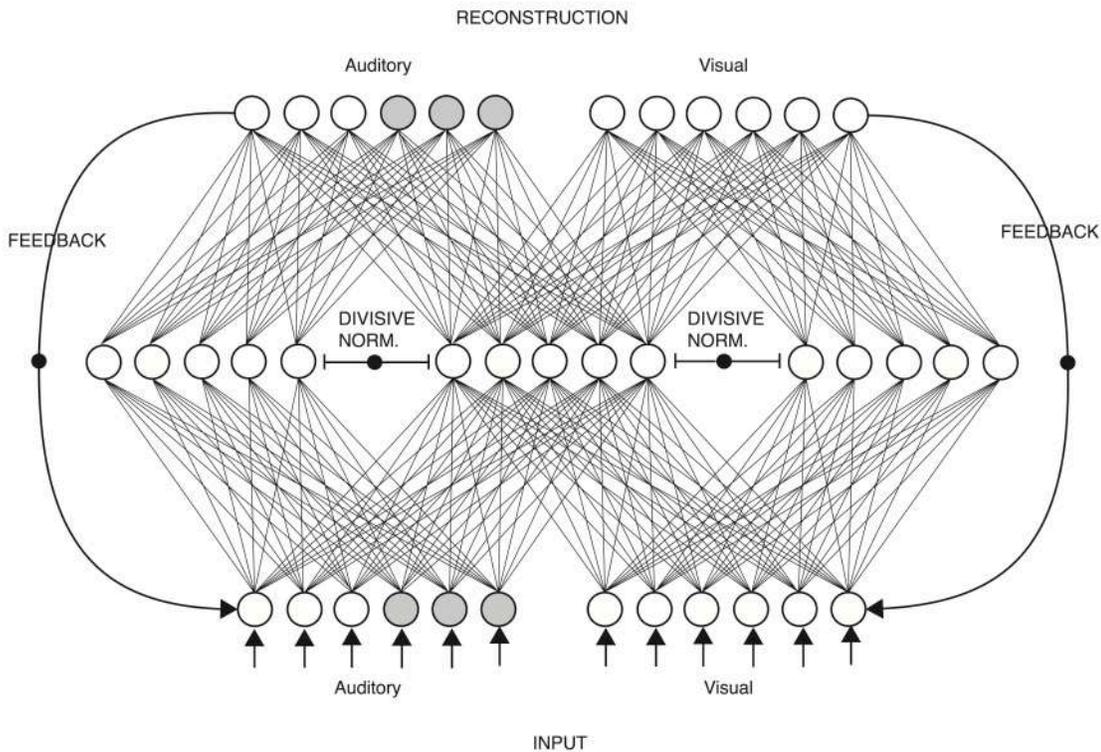

Fig. 1: Illustration of the network architecture showing the input, intermediate (pooling) and reconstruction layer (from bottom to top). Auditory input and reconstruction units are divided into subpopulations (e.g. unshaded units in the input and reconstruction layers could represent rightward-tuned units while shaded units could represent leftwards-tuned units).

### 2.1 The input layer

The input layer consists of two separate populations of unimodal units: a population of auditory units and a population of visual units. The auditory units are modeled after neurons in the auditory cortex (Werner-Reiss and Groh, 2008), as well as those observed in the SC (Lee & Groh, 2014), and are sigmoidally tuned to sound location (they have a rate code of spatial location). The auditory units are further subdivided into two subpopulations: one auditory subpopulation has responses that increase to the right side of the space, while the other subpopulation has responses that increase to the left side of the space; these two subpopulations could represent the neurons that are dominantly tuned to the contralateral side of the space, as found in the auditory cortices of both hemispheres (Werner-Reiss & Groh, 2008). The visual cortical neurons, with its units topographically tuned to the location of visual stimuli, follow a Gaussian profile (a tuning property also seen in visual SC neurons (Lee & Groh, 2014)). In the next section we will introduce the equations that describe the behavior of these subpopulations of units.

### 2.1.1 Auditory input units

The activity of each auditory input unit, tuned to the left side of space, $\theta_i^L$, for a given auditory stimulus position, $S_A$, is determined by the following equation:

$$\theta_i^L = \frac{\alpha_i^L g^A}{1 + exp\left(\frac{(S_A - x_i)}{m}\right)} \qquad , \tag{1}$$

where $\theta_i^L$ is the activity of unit $i$, where $\alpha_i$ is the adaptation (input) weight of the unit, which is set to 1 by default (unless adaptation is being modeled), where $g^A$ is the gain parameter, which together with $\alpha_i$ sets the maximum expected activity of the neuron, where $m$ is the rise parameter of the sigmoidal tuning function, which sets the rate at which the activity changes for a change in stimulus position. Finally, each unit's tuning function is centered at a specific location $x_i$ (that results in exactly half of the total expected gain).

Similarly, the activity of each auditory input unit tuned to the right side of space, $\theta_i^R$, for a given auditory stimulus position, $S_A$, is determined by the following equation:

$$\theta_i^R = \frac{\alpha_i^R g^A}{1 + exp\left(\frac{-(S_A - x_i)}{m}\right)} \qquad . \tag{2}$$

The parameters in the above equations are identical in both sub-populations, with exception of the $\alpha_i$ parameter, which is different for each unit tuned to each side of space when the adaptation is being modeled. Otherwise, by default all $\alpha_i$ are held constant at 1. The superscripts L and R, therefore, refer to the subsets of units tuned to the left and right, respectively. The superscript A represents the auditory sensory modality.

### 2.1.2 Visual input units

The activity of each visual input unit, $\theta_i^V$, given the visual stimulus positon, $S_V$, is described by the following equation:

$$\theta_i^V = g^V exp\left(\frac{-(S_V - x_i)^2}{2\sigma^2}\right) \qquad . \tag{3}$$

As above, the $g^V$ parameter sets the gain of the visual units. Each unit $i$ is maximally responsive to a stimulus at a given location $x_i$, and $\sigma^2$ is the variance of the Gaussian tuning profile (which adjusts the width of the tuning curve).

When modeling extremely wide tuning functions (large $\sigma^2$), there is a risk of a significant portion of the input activity profile getting cut off at the boundaries of our input space, causing undesired boundary effects. To prevent such boundary effects, we applied the following rule to the difference, $S_V - x_i$, in the numerator of the exponential in *Eq. 3*.

$$S_V - x_i = \begin{cases} |S_V - x_i|, & if\ |S_V - x_i| < L/2 \\ L - |S_V - x_i|, & if\ |S_V - x_i| > L/2 \end{cases},$$

where L is the width of the spatial locations considered (in our case, L = 301, since we modeled units tuned from -150 to 150 degrees in order to minimize the occurrence of boundary effects).

## 2.2   The intermediate pooling layer

The pooling (or intermediate) layer consists of three populations of units, inspired by the types of neurons found in the IP (Mullette-Gilman et al., 2005): one population contains unisensory auditory units, one population contains unisensory visual units and one population contains multisensory units. In the next section we will introduce the equations that describe the behavior of the units in each of the pooling layer populations.

### 2.2.1   The unisensory auditory pooling units

The activity of the unisensory auditory units of the pooling layer, $r_j^A$, given by its inputs $\theta_i^L$ and $\theta_i^R$ (as detailed above) is described by the following equations:

$$v_j^A = \sum_i w_{i,j}^L\ \theta_i^L + \sum_i w_{i,j}^R\ \theta_i^R \qquad , \qquad\qquad (4)$$

$$r_j^A = exp\left(v_j^A\right) \qquad ,$$

where $w_{i,j}^L$ is the weight of the $i^{th}$, leftward-tuned, auditory input unit on to the j$^{th}$ auditory pooling unit and $w_{i,j}^R$ is the weight of the $i^{th}$, rightward-tuned, auditory input unit on to the $j^{th}$ auditory pooling unit. $v_j^A$ represents the membrane potential of pooling unit j, which is a linear combination of the inputs multiplied by their respective weights. The activity of unit j, $r_j^A$, is finally obtained by applying a non-linear, exponential transformation to the membrane potential proxy, $v_j^A$.

The weights, $w_{i,j}^L$, of the leftward-tuned input units are sigmoidally distributed onto the pooling units according to the following relationship:

$$w_{i,j}^L = \frac{A}{n_L\left(1 + exp\left(\frac{-(x_i - x_j)}{m}\right)\right)} \qquad , \qquad\qquad (5)$$

where A is an auditory weight-scaling factor, $n_L$ is the number of units in the leftward-tuned input subpopulation, $x_i$ is the tuning curve mid-point of the input unit (as described above in the "auditory input units" section) and $x_j$ is the preferred position of the pooling neuron, where $m$ is the rate of weight increase across the population of units.

Similarly, the weights, $w_{i,j}^R$, or the rightward-tuned input units are distributed onto the pooling units according to the following relationship:

$$w_{i,j}^R = \frac{A}{n_R\left(1 + exp\left(\frac{(x_i - x_j)}{m}\right)\right)} \qquad . \qquad\qquad (6)$$

The parameters are the same as for the leftward-tuned inputs, except that $n_R$ is the number of units in the rightward-tuned input subpopulation. One can see that the only difference in the above two equations is the sign in the exponent. Overall, the distribution of the weights follow a sigmoid relationship opposite in direction to the tuning of the input units. This pattern of weight distributions give rise to Gaussian-like tuning properties in the pooling units with peak activity at their preferred locations.

### 2.2.2 The unisensory visual pooling units

The activity of the unisensory visual units of the pooling layer, $r_j^V$, given by its inputs $\theta_i^V$ is described by the following equations:

$$v_j^V = \sum_i w_{i,j}^V \, \theta_i^V \qquad , \qquad (7)$$

$$r_j^V = \exp\left(v_j^V\right) .$$

The membrane potential proxy, $v_j^V$, of the visual pooling units is a linear combination of its inputs $\theta_i^V$ multiplied by their weights, $w_{i,j}^V$. The activity of unit j, $r_j^V$, is obtained by applying a non-linear, exponential transformation to the membrane potential proxy, $v_j^V$.

The weights from the inputs to the pooling units follow a Gaussian distribution:

$$w_{i,j}^V = \frac{V}{\sigma\sqrt{2\pi}} \exp\left(\frac{-(x_i - x_j)^2}{2\sigma^2}\right) \qquad , \qquad (8)$$

where $x_i$ is the preferred location of input unit *i*, and $x_j$ is the preferred location of pooling neuron *j*. V is the visual weight-scaling factor. The parameter $\sigma$ is the width of the weight distribution function. The same rule for the distance in the numerator of the exponential in *Eq. 3* also applies here for $x_i - x_j$ to avoid boundary effects.

### 2.2.3 The multisensory pooling units

The activity of the multisensory units of the pooling layer, $r_j^m$, given by its inputs $\theta_i^V$, $\theta_i^L$ and $\theta_i^R$, is described by the following equations:

$$v_j^m = \sum_i w_{i,j}^{M,V} \, \theta_i^V + \sum_i w_{i,j}^{M,L} \, \theta_i^L + \sum_i w_{i,j}^{M,R} \, \theta_i^R + \mu , \qquad (9)$$
$$r_j^m = \exp\left(v_j^m\right) ,$$

The membrane potential proxy, $v_j^m$, of the multisensory pooling units are a linear combination of its inputs $\theta_i^R$, $\theta_i^L$ and $\theta_i^V$ multiplied by their weights , $w_{i,j}^R$, $w_{i,j}^L$, and $w_{i,j}^V$ respectively with the addition of an initial input (or bias) term, $\mu$, which represents input to the multisensory units from an external network which encodes the probability that the audio and visual stimuli have a common cause. How $\mu$ is determined will be covered in a later section (3.5). The activity of unit *j*, $r_j^m$, is obtained by applying a non-linear, exponential transformation to the membrane potential proxy, $v_j^m$.

The weights from visual input units onto the pooling units follow a Gaussian distribution:

$$w_{i,j}^{M,V} = \frac{V_m}{\sigma\sqrt{2\pi}} exp\left(\frac{-(x_i - x_j)^2}{2\sigma^2}\right) \qquad . \qquad (10)$$

The weights from auditory input units (both leftward tuned and rightward tuned) onto the pooling units follow sigmoidal functions:

$$w_{i,j}^{M,L} = \frac{A_m}{n_L\left(1+exp\left(\frac{-(x_i-x_j)}{m}\right)\right)} \qquad , \qquad (11)$$

$$w_{i,j}^{M,R} = \frac{A_m}{n_R\left(1+exp\left(\frac{(x_i-x_j)}{m}\right)\right)} \quad .$$

All parameters in *Eq. 10* and *11* are identical to those described in the unisensory pooling units (see subsections 2.2.1 and 2.2.2, above), except for $V_m$ and $A_m$ which are the multisensory auditory and visual weight scaling parameters, which, unless otherwise stated, are held constant at $V_m = 2$ and $A_m = 1$.

### 2.2.4  Divisive Normalization

The activities, $r$, of all units in the pooling layer are subject to divisive normalization following activation (Ohshiro et al, 2011). Each unit's activity in the pooling layer is normalized by the total average activity in the layer, where $n_V$ is the total number units in the pooling layer visual units-

$$r_j = \frac{r_j}{1+\left(\frac{\Sigma\, r_j^m + \Sigma\, r_j^A + \Sigma\, r_j^V}{n_R + n_L + n_V}\right)} \qquad (12)$$

Divisive normalization is used to keep activity in the intermediate/pooling layer to a minimum.

### 2.3  The reconstruction layer

The reconstruction layer consists of an auditory reconstruction subpopulation and a visual reconstruction subpopulation from the output of the intermediate pooling layer. Below, we describe how the reconstructed input is computed in i) the auditory reconstruction subpopulation and ii) the visual reconstruction subpopulation.

**Auditory Reconstruction Units**

The auditory reconstruction units sum up the activity of the auditory pooling units and multisensory pooling units, multiplied by the same weights used to compute the activity in the auditory pooling units and multisensory units. The reconstructed activity of the leftward and rightward auditory units, $\rho_i^L$ and $\rho_i^R$, are thus computed as follows:

$$\rho_i^L = \sum_j w_{j,i}^L\, r_j^L + \sum_j w_{j,i}^{M,L}\, r_j^M \qquad , \qquad (13)$$

$$\rho_i^R = \sum_j w_{j,i}^R\, r_j^R + \sum_j w_{j,i}^{M,R}\, r_j^M \quad .$$

**Visual Reconstruction Units**

Similarly, the visual reconstruction units sum up the activity of the visual pooling units and multisensory pooling units, multiplied by the same weights used to compute the activity in the visual pooling units and multisensory units:

$$\rho_i^V = \sum_j w_{j,i}^V\, r_j^V + \sum_j w_{j,i}^{M,V}\, r_j^M \qquad (14)$$

**Decoding stimulus location from the reconstructed input**

Assuming noiseless input and unit receptive fields centered at each possible location in space, the true location of auditory and visual stimuli can simply be read out by finding the auditory unit with half-maximal activity (due to sigmoidal tuning) and the visual unit with maximal activity (due to Gaussian tuning). When the input is noisy, however, one cannot simply use this heuristic on the input units to find the true location of stimuli. Nevertheless, since the reconstruction layer successfully produces a "noiseless" and smoothed reconstruction of the input, with the same number of units as the input layer and with units also centered at each location in space, one can simply read out the best estimate of stimuli location by using the heuristic described:

$$\hat{S}_V = argmax(\rho^V) \qquad , \qquad (15)$$

$$\hat{S}_A = argmax(\rho^R \otimes \rho^L) \; .$$

The auditory reconstruction unit with half maximum activity is equivalent to the maximum index of the element-wise product of reconstructed activities for right-ward and left-ward tuned units

## 2.4    Modeling recalibration of inputs

We modeled recalibration by dynamically adjusting the input adaptation weights, $\alpha_i$, according to the reconstruction error (the difference between the reconstructed input, $\rho_i$, and the input, $\theta_i$). If there is a stimulus at current time, t:

$$\alpha_i(t+1) = \alpha_i(t) + \eta \frac{\theta_i}{\max(\theta)} \left( \frac{\rho_i}{\max(\rho)} - \frac{\theta_i}{\max(\theta)} \right) \qquad (16)$$

where $\theta$ are the input activities, $\rho$ are the reconstructed activities and $\eta$ is the adaptation rate. If there is no stimulus at current time t, the weight decays back to the initial value of 1 at a fixed rate $\tau$:

$$\alpha_i(t=0) = 1$$

$$\alpha_i(t) = \begin{cases} \alpha_i(t-1) - \tau, \; \text{if } \alpha_i(t) > 1 \\ \alpha_i(t-1) + \tau \,, \text{if } \alpha_i(t) < 1 \\ \;\; \alpha_i(t-1), \; \text{if } \alpha_i(t) = 1 \end{cases}$$

# 3.  Model behavior and experimental results

## 3.1 Representation of unisensory estimates and reliabilities by the unisensory pooling units

To demonstrate the behavior of the model, we will first examine the roles of the pooling units in the intermediate layer. Each population of unisensory pooling units pools the activity (takes a weighted sum) of the input units of the appropriate sensory modality to generate a profile of activity. This profile of activity, normalized by the maximum value of activity, is fit almost perfectly by normalized Gaussian with the same peak and width (RMSE < 0.01, see *fig. 2*). Therefore, we take the peak of this profile of activity to represent the maximum likelihood estimate for the unisensory stimulus location and the width of the profile of activity (approximately 2.355 multiplied by the width at half-maximum) to represent the standard deviation of a Gaussian likelihood function. Indeed, when noiseless input is given to the unisensory populations of the intermediate layer, the peak of the generated activity profile is over the true location of the stimulus: the maximally activated unit has a receptive field center equal to the true stimulus location.

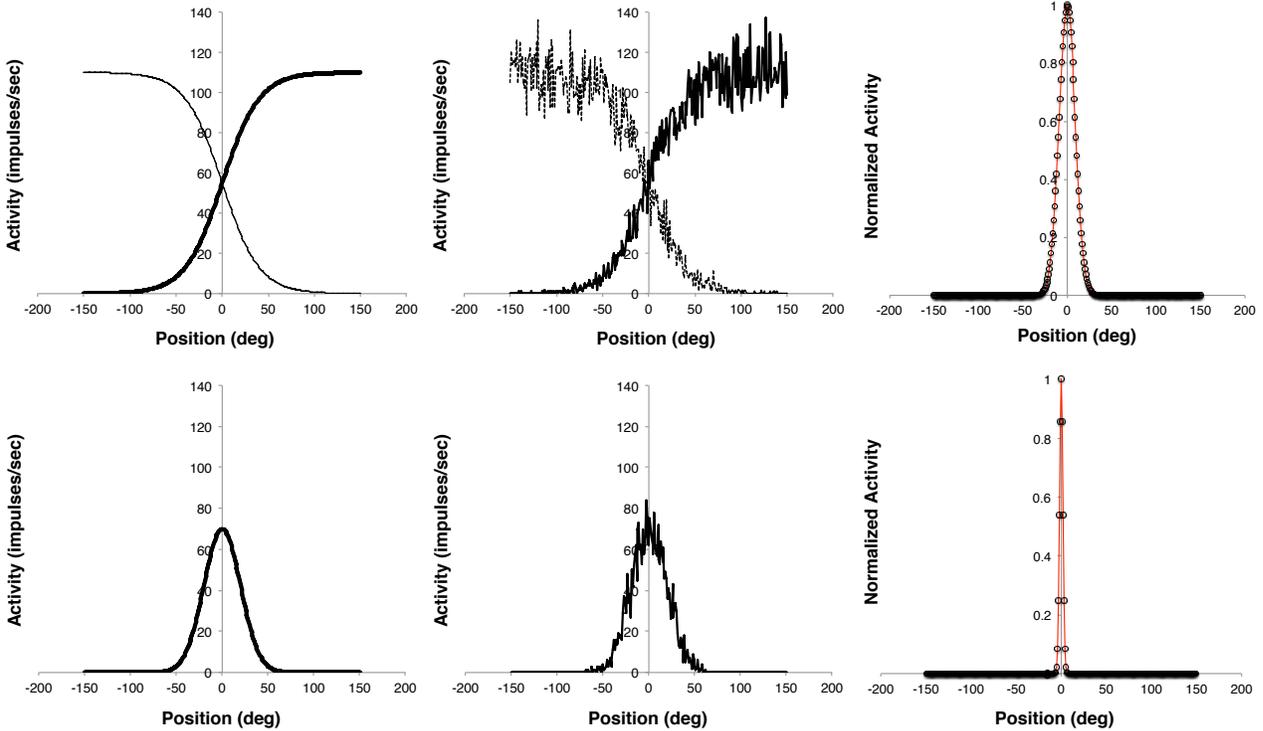

Fig. 2: A plot of activity in the input and unisensory pooling units; each unit's receptive field is centered over a particular region of space, in steps of 1 degree. The populations of unisensory pooling units for auditory (top-right) and visual (bottom-right) modalities generate a profile of activity (black points) that conforms to a Gaussian function (red solid line). The peak and widths of these activity profiles/Gaussians are taken to represent the unisensory maximum likelihood estimate and likelihood standard deviations (of which the squared reciprocal is equal to the unisensory reliability). Both auditory and visual stimuli are at position 0. Either noiseless (leftmost column) or noisy (center column) inputs will give rise to smooth Gaussians in the unisensory pooling units, peaked at $x_A$ and $x_V$ with standard deviations of $\sigma_A$ and $\sigma_V$. Top row: Auditory units (both leftward and rightward-tuned subpopulations are shown); Bottom row: Visual units.

We must note, however, that the abstract representations of likelihood functions generated by the unisensory pooling populations do not represent the likelihood distribution based solely on the input alone, as would be generated by an optimal probabilistic population code such as the one proposed by Ma and colleagues (Ma et al., 2006). The variance of the Gaussian profile produced by the unisensory pooling units is in fact larger than what is expected for an optimal probabilistic population code; we take this to reflect a suboptimal pooling of the input activity which gives rise to a greater uncertainty than what is expected from the input alone.

In the model, different unisensory reliabilities (the reciprocal of the likelihood function's variance) can be achieved by adjusting the gain or tuning widths of the input units. There is an inverse relationship between the gain of the input units and the width of the activity profile of the unisensory pooling unit: with a uniform increase in the gain of the input units, the width of the profile of activity (width of likelihood function) in the unisensory pooling units decrease non-linearly. Conversely, there is a direct linear relationship between the width of the input tuning curves (or rise parameter in auditory units) and the width of the activity profile of the unisensory pooling unit: a uniform increase in the tuning widths of the input units results in a proportional increase in the width of the activity profile (width of likelihood function) in the unisensory pooling units.

One appealing characteristic of the model is that the pooling layers produce smooth activity profiles even in the presence of input noise. When Poisson noise (variability that is equal to expected activity) is added to the input activity, the pooling layer is still able to represent a smooth likelihood function; although, one that varies slightly in peak and width from trial to trial. We use Poisson noise, since spike counts in a wide variety of cortical areas follow Poisson-like statistics (Shadlen & Newsome, 1998). The average peak of the activity profile and the average width of the activity profile over many repeated trials, however, still faithfully represent a maximum likelihood estimate and uncertainty when there is no input noise (see *fig. 2*).

### 3.2 Representation of the multisensory estimate and degree of causal relatedness by the multisensory pooling units

The multisensory pooling units pool the activity (take a weighted sum) of the input units from both sensory modalities to generate a population-wide profile of activity. The resulting profile of activity would peak approximately at the multisensory (full cue-combination) location estimate, if one uses the uni-sensory reliabilities and maximum likelihood estimates derived from the activity profiles of the unisensory pooling units (see *fig. 3*).

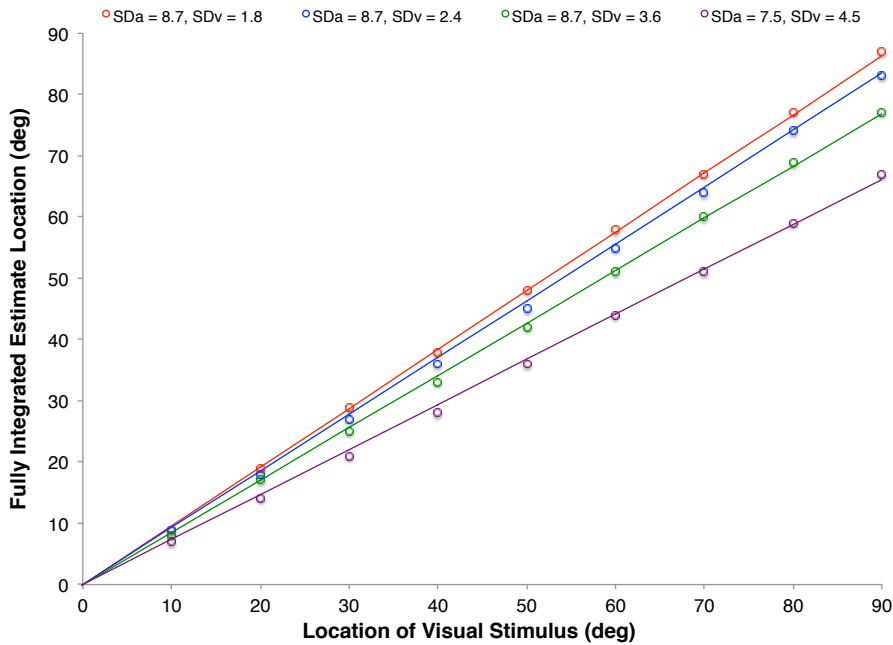

Fig. 3: The peak of activation in the intermediate multisensory population (open circles) closely follows the fully integrated estimate produced by optimal cue-combination (solid lines). The auditory stimulus was held at position = 0 deg, while the location of the visual stimulus was varied; input was noiseless. Different colors represent different input parameter settings (tuning curve widths) while input gains were held constant (visual gain = 70, auditory gain =120); this resulted in different unisensory reliabilities (see legend for unisensory SDs). Parameter settings: red: $m = 20$, $\sigma = 20$; blue: $m = 20$, $\sigma = 30$; green: $m = 20$, $\sigma = 40$; violet: $m = 15$, $\sigma = 50$.

The total activity of the intermediate multisensory population reflects the causal relatedness of auditory and visual stimuli. To illustrate this point, we compared the total activity of the multisensory population and the total activity of the auditory unisensory population, both as a percentage of their combined activity, during auditory and visual stimulation of varying spatial disparities. As the disparity between auditory and visual stimuli shrinks, there is a non-linear increase in the total activity of the multisensory population, while the total activity of the unisensory population remains constant. Therefore, the unisensory activity, as a percentage of the total activity combined in both populations, reaches its minimum when the multisensory activity is at its maximum; this occurs when the auditory and visual stimuli are at the same location. At sufficiently large separations between stimuli, the opposite is true: the multisensory activity dips down below the level of unisensory activity. This relationship between unisensory and multisensory activity nicely mirrors the relationship between the posterior probabilities for common and uncommon causes (see *fig. 4*).

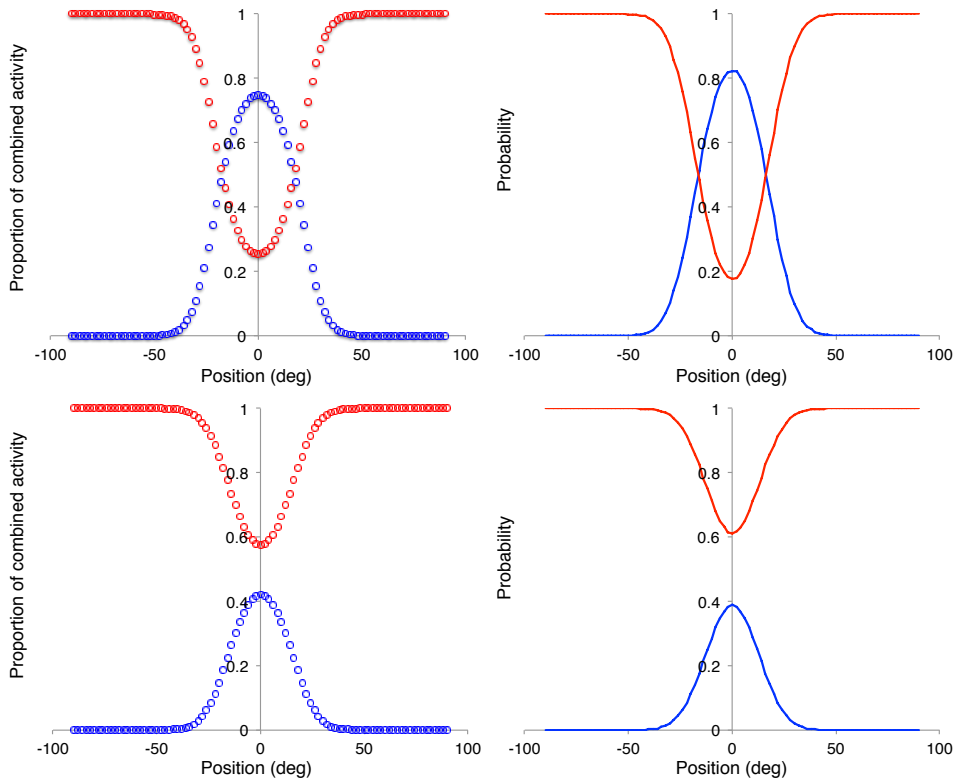

Fig 4: Top row: The plot on the left shows the total activity in the intermediate auditory unisensory (red) and multisensory (blue) populations as proportions of their total combined activity across different spatial disparities (the initial multisensory input, $\mu$, was set to 10.5). This pattern of activity across disparities mirrors the relationship between the posterior probability of the one-cause (C=1) and two-cause (C=2) models, as shown in the blue and red plots respectively, on the right ($p_{common}$ = 0.5). Bottom row: Same as in the top row, only with a smaller initial multisensory input, $\mu$, of 9.1 used to generate the plot on the left and a corresponding $p_{common}$ of 0.1 used to generate the plot on the right.

## 3.3 Initial input to the multisensory population represents a "prior" for causal relatedness

Given that the overall level of activity in the multisensory population mirrors the probability that two stimuli are related, based on the spatial alignment of inputs, the "initial input" term, $\mu$, that is added to the multisensory activity represents the prior that the stimuli are related. This initial input comes from a hypothetical external network that compares the relatedness of cues based on meaning, associations, context, or other factors orthogonal to spatial and temporal alignment. For now, we treat it as a free-parameter that scales the activity of the multisensory population, without changing the location of its peak (see *fig. 4*) and therefore influences the probability that two stimuli are related ($p_{common}$). In a later section (3.5) we characterize the approximate relationship between $p_{common}$ and $\mu$.

### 3.4 The reconstruction layer

Because the reconstruction layer pools inputs from both the intermediate multisensory and unisensory populations, the reconstructed activity will more closely reflect the dominant population's representation of stimulus location (full integration vs. full segregation) in a manner that is weighted by the activity of these populations. Therefore, the pooling of activity from the intermediate populations by the reconstruction units is analogous to model averaging in the causal inference model. To test this proposition, we carried out simulations in which we presented auditory and visual stimuli to the model, with varying spatial separations (the auditory position was held at 0, but the visual stimulus was varied from -90 to 90 degrees in steps of 2 degrees). We used the expected input (no noise), as determined by the input tuning curves, in the first layer to simulate overall expected behavior. The final estimates for auditory and visual stimulus locations were decoded from the reconstruction layer as described in *Eq. 15* (see also *fig. 5*). These values are compared to the mean estimates predicted by the Causal Inference model (Körding et al 2007), assuming a uniform prior over space, which were computed through Monte Carlo simulations: 10,000 $x_A$ and $x_v$ samples were drawn from Gaussians centered on the simulated stimulus locations, with widths of $\sigma_A$ and $\sigma_v$ (same values derived from the intermediate unisensory populations); these samples were then used as input into the Causal Inference model to generate final location estimates, based on model-averaging. For our simulations, we discretized the range of hypothesized stimulus locations between -90 and 90, in steps of 1 degree. With the initial multisensory input parameter, $\mu$, as a free parameter, the simulations described above produced excellent fits to the average predictions of the Causal Inference model for a variety of input settings and $p_{common}$ values; we will describe these simulations in more detail in the following section.

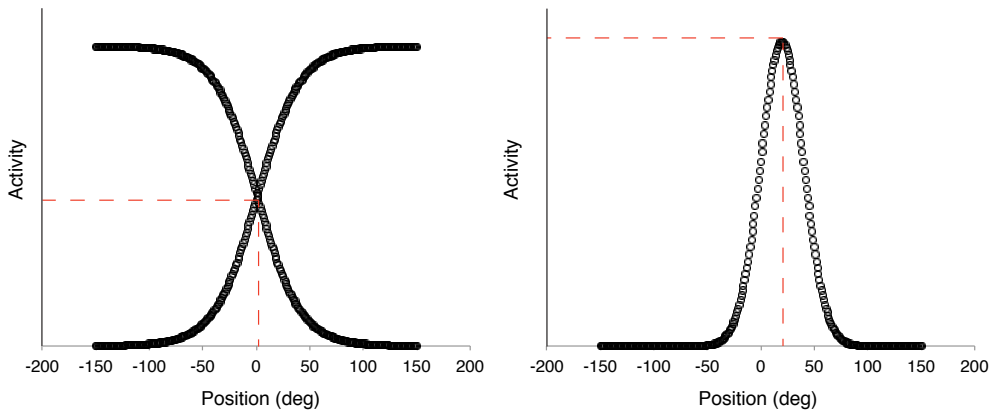

Figure 5: Activity in the populations of the reconstruction layer (left plot, auditory sub-populations; right plot, visual population). The estimate for the auditory stimulus location, $\hat{S}_A = 13$, is decoded as the receptive field position of the unit(s) whose activity is at half-maximal (red dotted line, left plot). The estimate for the visual stimulus location, $\hat{S}_V = 20$, is decoded as the receptive field position of the unit whose activity is at maximal (red dotted line, right plot). For this particular simulation the input parameters were $\sigma = 20$, m = 20, $g^A = 140$ $g^V = 80$; the initial multisensory activity was set to $\mu = 12.3$ which corresponded to a $p_{common} = 0.95$. True stimulus locations were at $S_A = 0$, and $S_V = 20$.

### 3.5 The relationship between $p_{common}$ and initial multisensory input, $\mu$

Let us first consider the situation in which the input parameters are fixed, i.e. the stimulus intensity or contrast remains constant, but the prior probability that two stimuli are related ($p_{common}$) varies, i.e. the semantic congruence of the stimuli changes. To model such a situation we ran simulations as described directly above while varying $p_{common}$. We then determined the best-fitting initial multisensory input, $\mu$ parameter (with a resolution of 0.05). Overall, the model produces excellent fits across different values of $p_{common}$ (see *fig 6*).

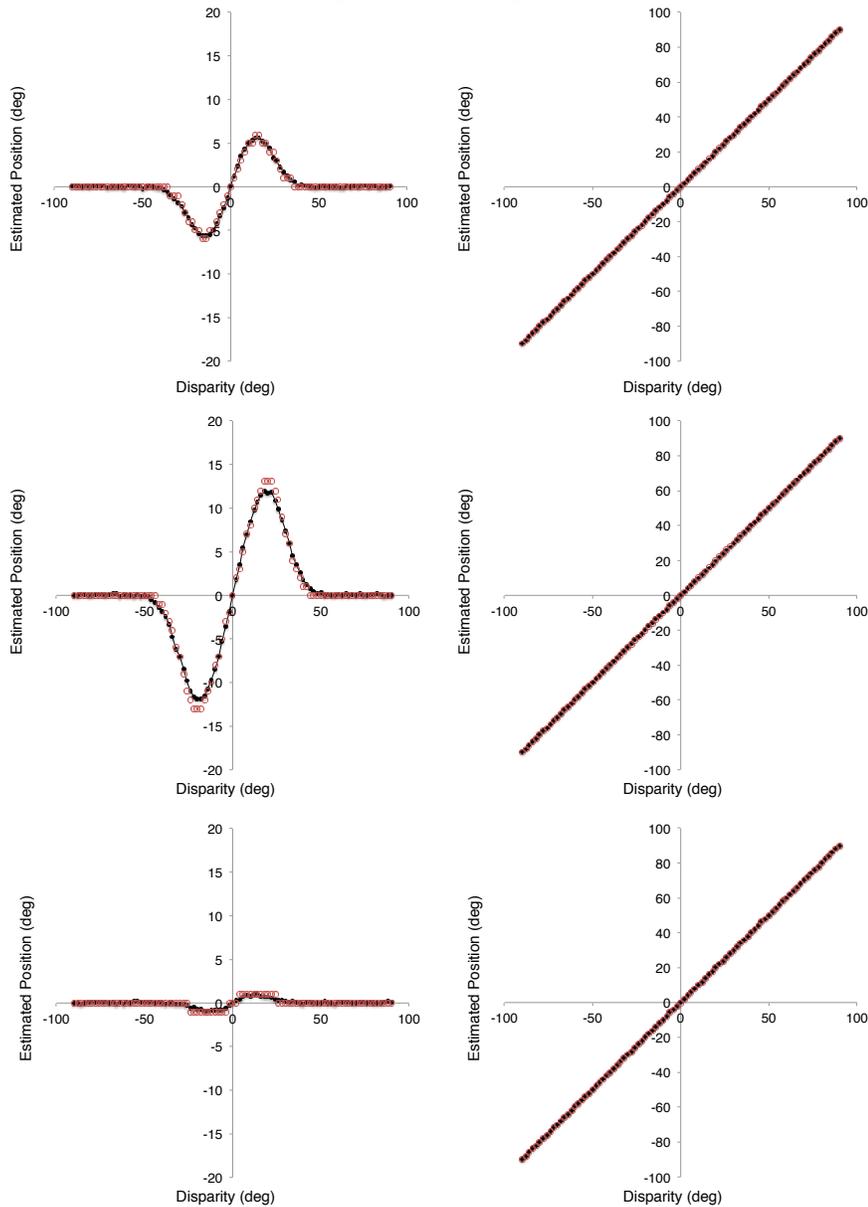

Figure 6: Left Column: Position estimate for auditory stimulus across disparities produced by the network (red open squares) gives a good fit to the mean auditory position estimates produced by the Causal Inference Model (black points connected by dotted-line). Right Column: Same as in the left column, only for visual position estimates. Input parameters were held constant: m = 20, σ = 20, $g^V$ = 80 and $g^A$ = 140 (SDa = 8.1. SDv = 1.7). Top row: Simulating $p_{common}$=0.5 by setting initial multisensory input, mu = 10.5 (left plot RMSE = 0.252, right plot RMSE =0.106). Middle row: Simulating $p_{common}$= 0.95 by setting initial multisensory input, mu = 12.3 (left plot, RMSE = 0.47; right plot, RMSE = 0.24). Bottom row: Simulating

$p_{common}$ = 0.05 by setting Initial multisensory input, mu = 8.65 (left plot, RMSE = 0.197; right plot, RMSE = 0.24).

By plotting the best-fit $\mu$ parameter as a function of $p_{common}$, an interesting pattern emerges (see *fig. 7, top*): the best-fit $\mu$ increases monotonically, but non-linearly, across $p_{common}$. This relationship is fit well by a scaled logit function of the following form:

$$\boldsymbol{\mu} \approx \left( \frac{\log\left(\frac{p_{common}}{1-p_{common}}\right)}{0.7} \right) + \boldsymbol{c} \qquad , \qquad\qquad (17)$$

where the constant, *c*, is the initial multisensory input when $p_{common}$ is 0.5, and the left-hand term becomes 0. We repeated these simulations for a number of different input gain ($g^A$ and $g^V$) parameter combinations to simulate different unisensory reliabilities, and the relationship described by *Eq. 17*, held true throughout (see *fig. 7, bottom*).

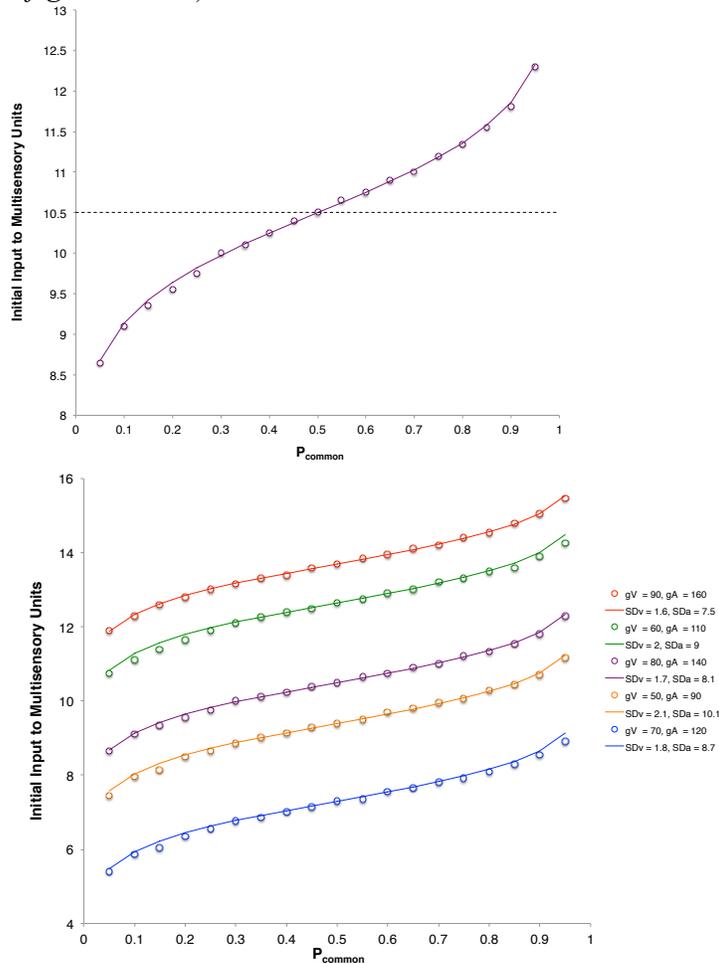

Figure 7: Top: Best-fit $\mu$ (initial multisensory input) for a single set of input parameters (m = 20, σ = 20, V = 80, A = 140) across different $P_{common}$ values (open circles) are well fit by *Eq. 17* (solid line), RMSE = 0.039. The dotted line represents the *c* parameter, which is the initial multisensory input when $p_{common}$ = 0.5. Bottom: The relationship holds for a number of tested input gains (see legend; tuning widths are held constant at m = 20, σ = 20). The average RMSE for the displayed curves is 0.064.

Interestingly, the constant, *c*, appears to depend on the level of input activity (set by gain parameters and tuning widths), even for a fixed $p_{common}$. Let us assume that an external network uses the same input units as the network described here. However, the external network is unique in that it pools the input (including other input units, we have not modeled here, tuned to different stimulus properties, such as sound frequency, shape or color) in order to generate output that represents the relatedness of the stimuli based on contextual cues. The output from the external network is composed of two components, and only one of which (the left-hand term in *Eq. 17*) truly depends on $p_{common}$, while the other component (the right-hand term) is a nuisance parameter that is independent of $p_{common}$ but still depends on the overall activity of the input units. The overall output from this hypothetical external network is used as initial input, $\mu$, to the multisensory units in the network described here. To characterize the relationship between the input gain and the *c* term required to fit the Causal Inference model, we ran simulations as described above with a number of combinations of auditory and visual input gains but only for constant $p_{common} = 0.5$ (see *fig. 7, bottom*). Overall, the model produced excellent fits to the mean causal inference estimates (see *fig. 8*). If we hold the visual gain parameter constant, while varying the auditory gain one can see that there is a positive linear relationship between the auditory gain parameter and the best-fit *c* (see *fig. 9, top*). Conversely, there is a negative linear relationship between the visual gain parameter and *c*. By combining these linear relationships, we can approximately predict the best-fit *c*:

$$c \approx 0.866 g^A - 1.4025 g^V + 1.616. \tag{18}$$

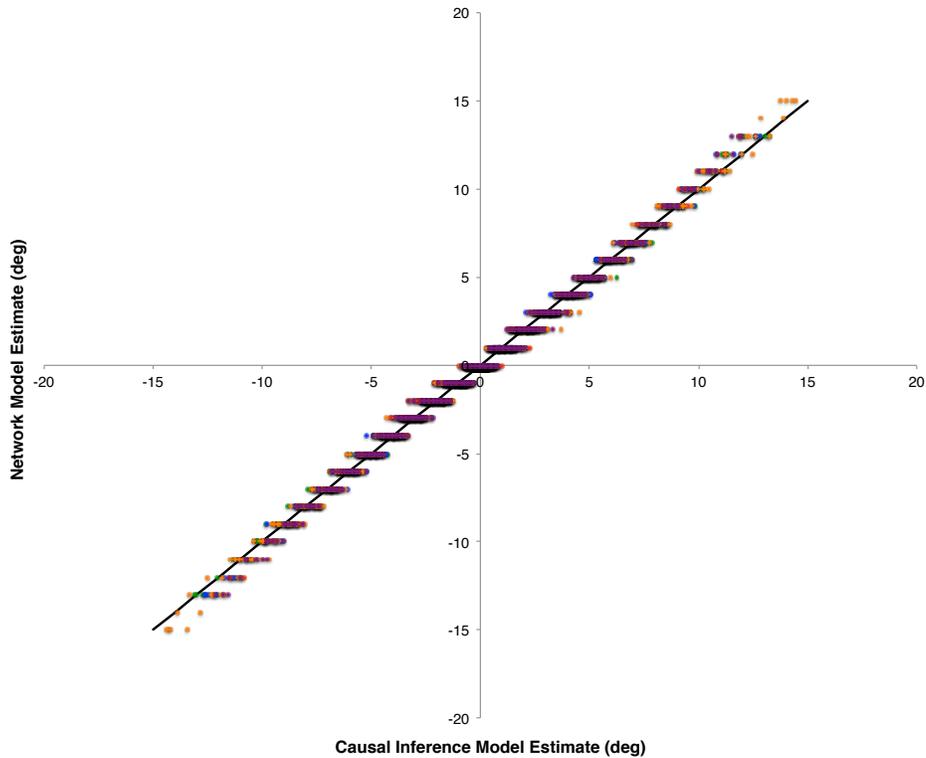

Figure 8. Causal inference model estimates plotted against best-fit network model estimates for a number of different unisensory reliability settings (same as in figure 7). See figure 7 legend for colour coding. $R^2 = 0.991$.

A surface plot shows the relationship between auditory gain, $g^A$, the visual gain, $g^V$, and the *c* parameter (see *fig. 9, bottom*).

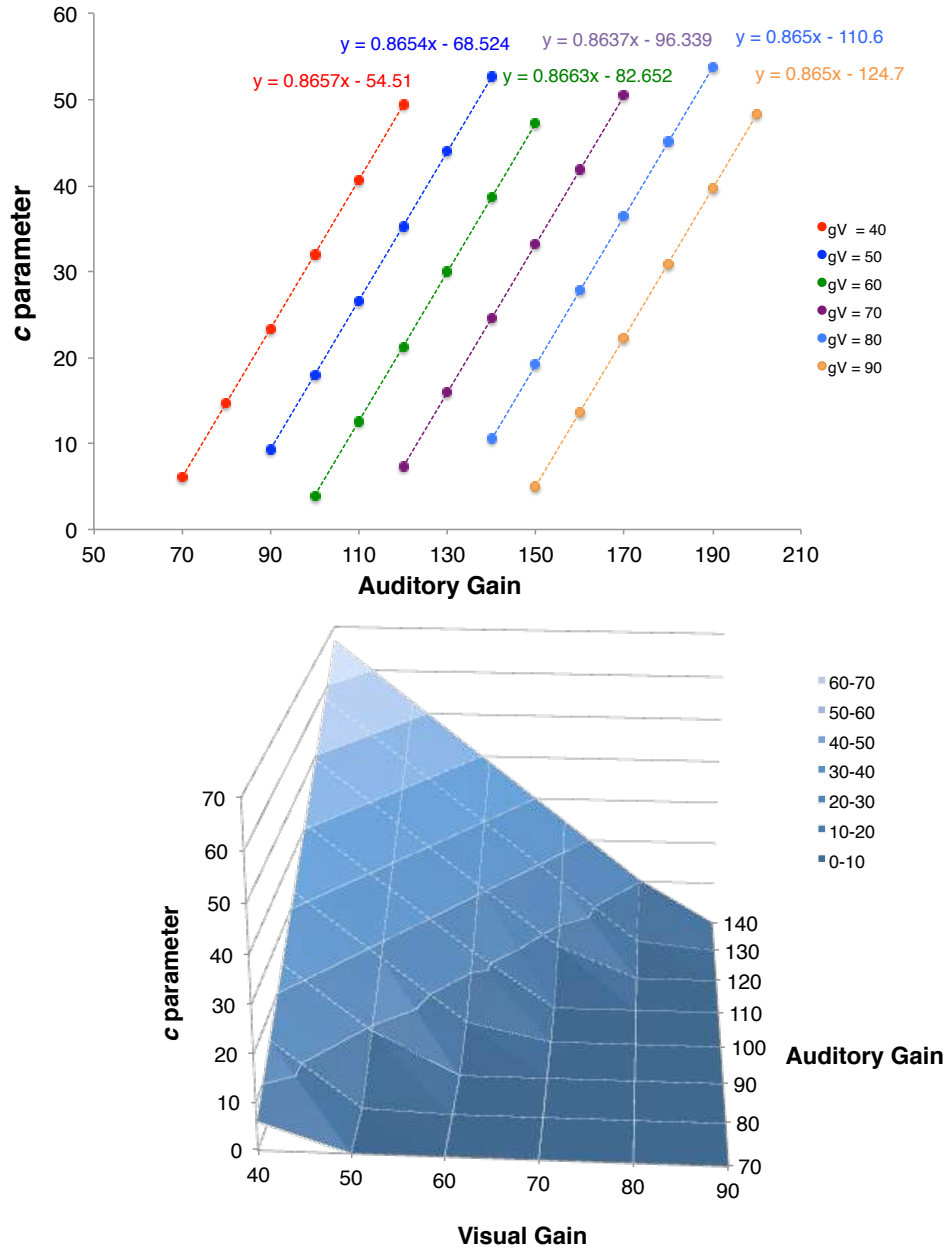

Fig 9: Top: The linear relationship between auditory gain, $g^A$, and the intial input to the multisensory layer necessary to model $p_{comon}$ =0.5, also known as the $c$ parameter, plotted for different visual gains, $g^V$.  Bottom: The relationship between auditory gain, visual gain and the $c$ parameter depicted as a surface plot, derived from the first plot.  The surface plot, in actuality, dips below 0 on the vertical axis, however for clarity we only show positive activity values.

We simulated a situation in which the reliability of visual and auditory stimuli are close to equal (auditory SD = 8.1 and visual SD = 7.5). In this case, there would be much less ventriloquism effect for the auditory stimuli than in the previously modeled cases, but the visual stimulus would experience more ventriloquism effect than before. We simulated this condition by using the same gain parameter settings as in *fig. 6*, but drastically increased the tuning width of the visual inputs (σ = 80); this also required us to slightly adjust the *V* parameter to 4.335.  Using the relationships above to determine the required initial input to the multisensory units, we ran simulations across different $p_{common}$ values (same simulation methods as in *fig. 6*) and generated good fits to the causal inference model (see *fig. 10*)

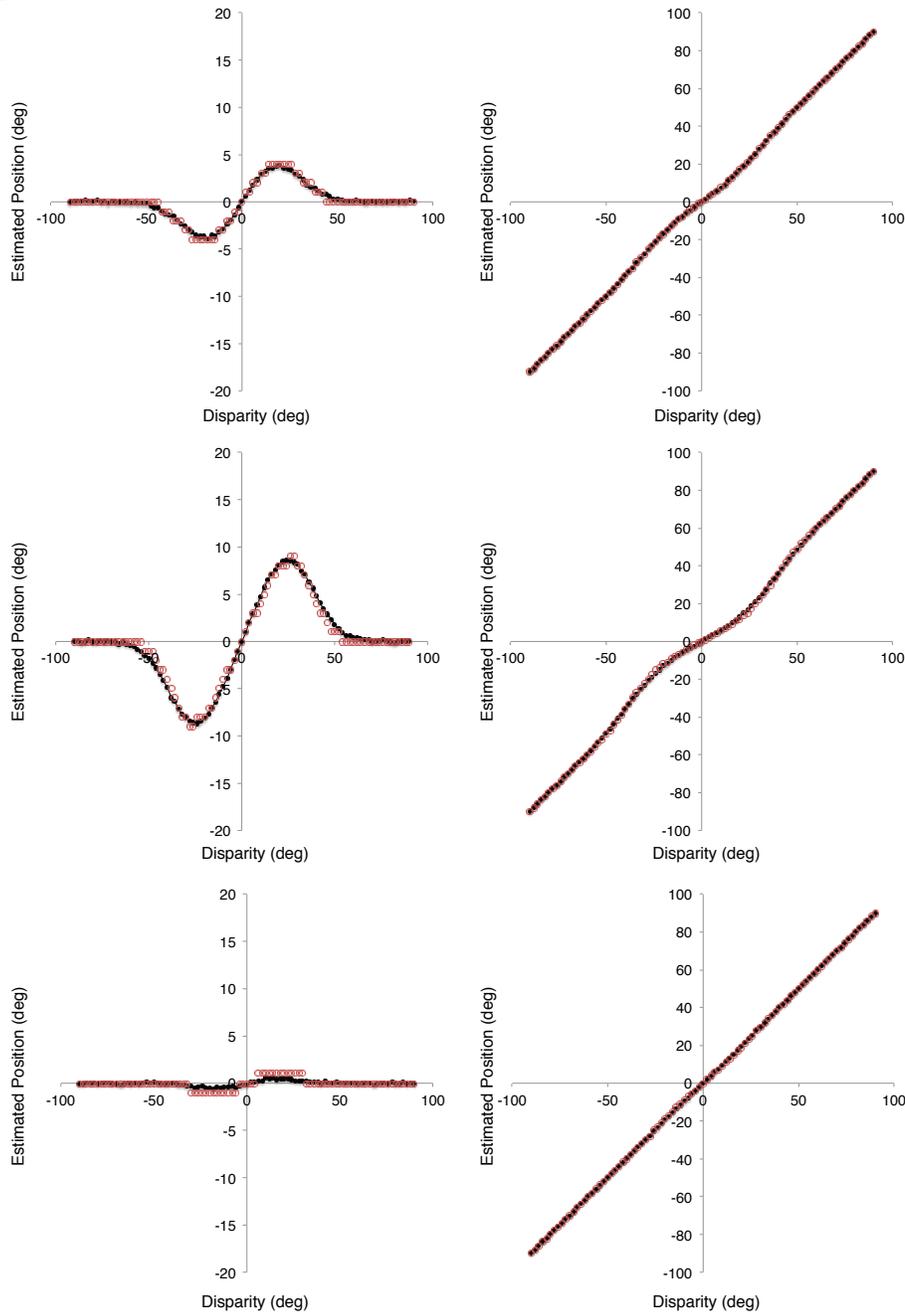

Fig 10: Left Column: Position estimate for auditory stimulus across disparities produced by the network (red open squares) gives a good fit to the mean auditory position estimates produced by the causal inference model (black points connected by dotted-line). Right Column: Position estimate for visual stimulus across disparities produced by the network (red open squares) gives a good fit to the mean visual position estimates produced by the causal inference model (black points connected by dotted-line). Input parameters were held constant: m = 20, σ = 80, $g^V$ = 80 and $g^A$ = 140 (SDa = 8.1. SDv = 7.5). Top row: Simulating $p_{common}$=0.5 by setting initial multisensory input, mu = 10.5 (left plot RMSE = 0.29, right plot RMSE =0.24). Middle row: Simulating $p_{common}$= 0.95 by setting initial multisensory input, mu = 12.3 (left plot, RMSE = 0.52; right plot, RMSE = 0.64). Bottom row: Simulating $p_{common}$= 0.05 by setting initial multisensory input, mu = 8.65 (left plot, RMSE = 0.31; right plot, RMSE = 0.29).

**3.6    The calibration of input units by feedback of reconstructed activity**

As demonstrated in the previous section, the reconstructed activity from both auditory and visual populations of the reconstruction layer encodes the final location estimate(s) produced by the causal inference model. This activity could be fed forward to decision making and/or motor planning networks, but the activity could also be fed back to lower layers of the processing hierarchy for prediction or, for the purpose we model here, recalibration. Because the final location estimate is regarded as a more informed estimate than the one produced by earlier processing stages (the likelihood estimate), as it incorporates information from more than one sensory system in light of its generative models, the reconstructed activity is used by our network as the "template" against which the input is compared. The error, or difference, between the profile of reconstructed activity and the original input activity is computed, and the input weights are readjusted accordingly (see *Eq. 16*). Therefore, the greater the shift in final causal inference estimate, compared to the unisensory maximum-likelihood estimate, (i.e. the ventriloquism effect) the greater the overall "error" produced by the input units and the more adjusted the weights.

After weight readjustment, one can simulate the ventriloquism aftereffect by giving a unisensory auditory stimulus and decoding, from the reconstruction layer, the final estimate for auditory location. We found the combination of recalibration parameters, $\eta$ (the adaptation rate) and $\tau$ (the decay rate), that provided the best fits to data from Bosen et al (2007). In their experiment, Bosen and colleagues presented trains (20 presentations) of simultaneous audio (noise bursts) and visual (laser point) stimuli with a constant spatial disparity (8 degrees). Each successive presentation was separated by approximately 1 second. They either measured participants' localization estimates for the sound stimulus that occurred with a visual stimulus (during the train), or for a subsequent auditory stimulus that happened alone, either 1, 5 or 20 seconds following the stimulus train. The former measurement reflects the online ventriloquism effect, while the latter measurement reflects the ventriloquism aftereffect and the time course of its decay. We simulated both tasks by first assuming that each trial duration is 1 second long; we gave the network 20 simultaneous auditory and visual stimuli interleaved with single trials with no stimulation (letting the adaptation weights decay for 1 second). In all of these trials we set the auditory stimulus at $S_A = 0$ deg, and the visual stimulus at $S_V = 8$ degrees; then we gave an additional unisensory auditory stimulus ($S_A = 0$ deg) at 1, 5 or 20 seconds following the stimulus train. The model gave a final auditory location estimate for each stimulus trial. For the simulation of localizing the auditory stimulus when presented with the discrepant visual stimulus (within the stimulus train), there was a clear and relatively strong ventriloquism effect that remained constant at each stimulus presentation (mean auditory localization ≈ 5 deg); the model replicated this result with input parameters set to $m = 20$, $\sigma = 20$, $g^A = 140$, $g^V = 80$, and $\mu = 10.7$ (corresponding to a $p_{common} \approx 0.57$). For the same parameter settings, the model gave a good fit to the ventriloquism aftereffect and the time course of its decay (RMSE = 0.43) when the adaptation parameters were set to $\eta = 0.65$ and $\tau = 0.009$ (see *fig. 11, left*).

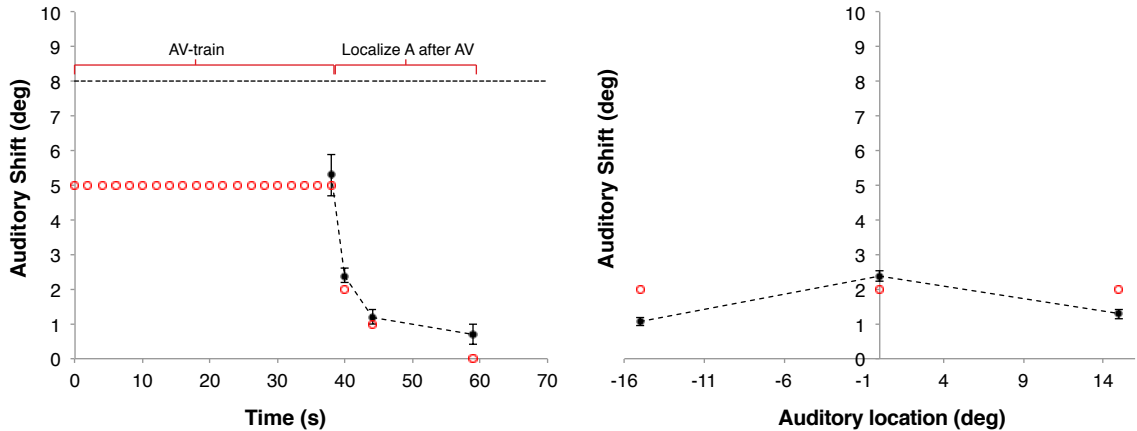

Figure 11: **The time course of the Ventriloquism Aftereffect (left)**: The first 20 trials are simulations of the AV (auditory and visual) train and the localization of the auditory stimulus (A) during these AV presentations. The last 3 trials are separate simulations in which A was localized 1 second, 5 seconds and 20 seconds after the end of the AV train. The location estimates produced by the model (red open circles) fit well with the data from Bosen et al. (2007) which demonstrates visual capture, the ventriloquism aftereffect and the time course of its decay (black points, the error bars represent ± 1 SE). The dotted line at 8 degrees indicates the constant AV disparity.
**Extension of the Aftereffect to local space (right):** The presentation of an auditory stimulus 1 second following an AV train results in an auditory shift, regardless of whether the A occurs at the same location as the adaptation stimulus (0 deg) or ±15 deg from the adaptation stimulus. The auditory shift predicted by the model (with identical parameter settings as the plot on the left) is comparable to that which was measured by Bosen et al. (2007).

Bosen et al. (2007) further tested the stability of the online ventriloquism effect after many AV presentations by comparing the localization of the auditory stimulus of an AV pair after one AV presentation against the localization of the auditory stimulus of an AV pair after 20 AV presentations. The ventriloquism effect remained unaffected by the train of AV stimulation. Conversely, the ventriloquism aftereffect, measured when participants localized a unisensory auditory stimulus after 1 or 20 AV repititions, was significantly increased by the AV stimulus train. We simulated these experiments with our network model as described in the previous paragraph, with the same parameter settings, except for the initial input to multisensory units, which we changed to μ = 11.3 to fit the slightly higher auditory shift that was measured in this set of experiments. Figure 12 shows the results: our network overall predicted the trends reported by Bosen et al (2007) quite successfully.

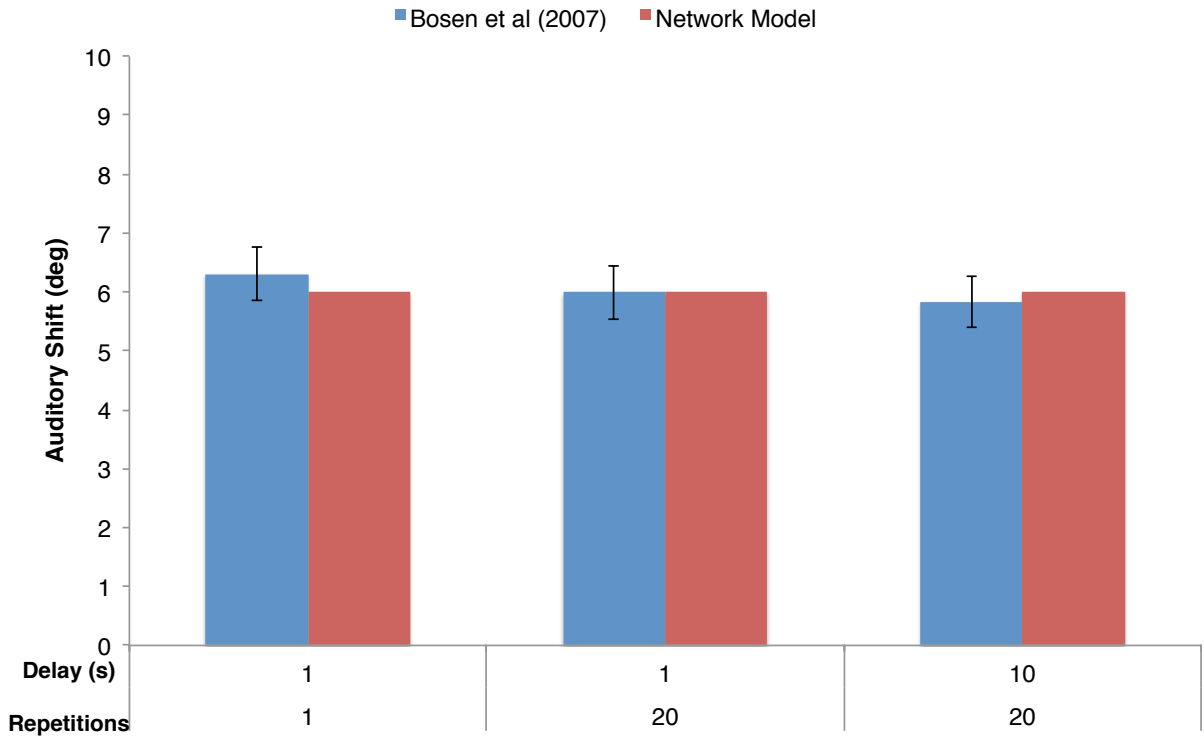

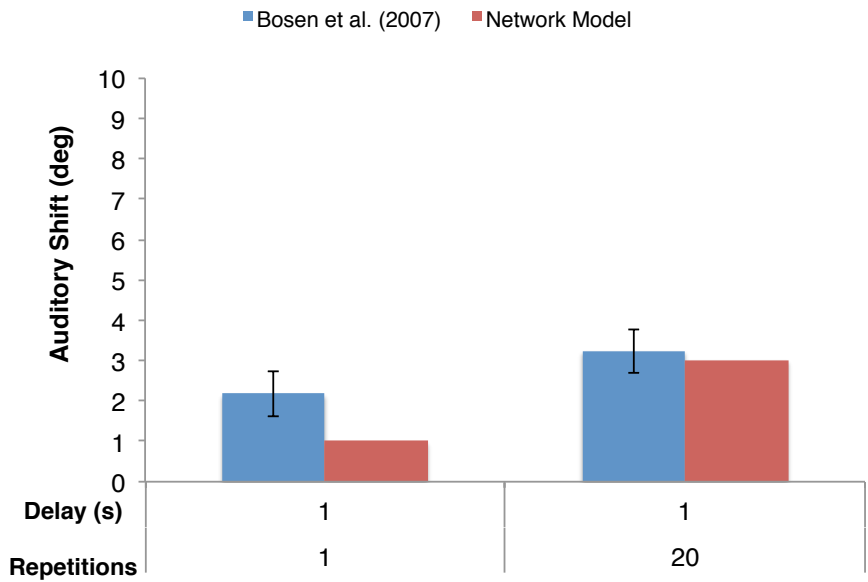

Figure 12: Simulations of experiment 2 from Bosen et al. (2007). Averaged data taken from study in blue (error bars show ±1 SE), model simulation results in red. The AV disparity was held constant at 8 deg. The auditory shift measured during the localization of the auditory stimulus within AV presentations (online ventriloquism effect) for different numbers of AV repetitions and delays is shown in the top plot. The auditory shift measured following AV presentations (ventriloquism aftereffect with a delay of 1 second) for different numbers of repetitions is shown in the bottom plot.

## 4. Discussion

We have presented a computational model that successfully approximates a wide range of predictions and estimates produced by the expected behavior of the causal inference model, which in turn captures the behaviour of observers in sound-localization experiments in the presence of simultaneous visual stimuli (the ventriloquism effect and its breakdown for large spatial disparities). Additionally, we have shown that the model can also produce good fits to human behavior in the ventriloquism aftereffect: our model reproduces the immediate and cumulative effects of the ventriloquism aftereffect as well as the time course of its decay with a singular set of adaptation parameters.

## 4.1 Parallels between model behavior and electrophysiological and neuroimaging findings

The neuroimaging and EEG studies reviewed in the introduction section (1.2) have greatly inspired this work. Now that we have given a detailed description of the model's behavior, we wish to revisit the findings of some of these studies to highlight what we consider preliminary support for our proposed architecture and dynamics.

According to a study by Rohe and Noppeney (2015), different brain regions along the dorsal processing hierarchy show activation patterns that correlate with the estimates produced by the causal inference model. Each distinct population of units in our model also produces activity that uniquely correlates with particular estimates in the causal inference model. For instance, in the input layer, auditory and visual neurons produce activity that correlates directly with the locations of auditory and visual stimuli, respectively. The total activity of left-ward tuned auditory input neurons, relative to that of the rightward tuned auditory input neurons, correlates with the azimuthal sound location. Since the Planum Temporale (PT) of each hemisphere contain neurons dominantly tuned to the contralateral side of space (Werner-Reiss & Groh, 2008), the activation pattern produced between left and right auditory cortical areas should match the types of activation patterns found by Bonath et al. (2007) and Zierul et al (2017), in which the activation of PT is correlated with how far to the contralateral side the stimulus is perceived to be. This also fits with the finding from Rohe and Noppeney (2015) that auditory cortex and higher auditory cortical areas (PT) have activation patterns that correlate with the unisensory estimate of sound location ($\hat{S}_{A,c=2}$). Since visual inputs (both in this model and in visual cortex) are retinotopically aligned and utilize a place code for spatial location, they would also produce activity that directly correlates with the visual stimulus location.

The population of multisensory units in the intermediate layer produces a peak of activation that represents the location of the forced fusion estimate and it's overall level of activation represents how probable the cues are to be related. The activation of posterior intraparietal sulcus (IP1-2) was also found to correlate with the estimate for the forced fusion estimate ($\hat{S}_{c=1}$) (Rohe & Noppeney, 2015). The intermediate unisensory units of our model also produce estimates for the segregated unisensory estimates of sound location ($\hat{S}_{A,c=2}$ and $\hat{S}_{V,c=2}$) by creating peaks at these locations, however, the overall level of activity of these units is independent of the spatial disparity and probability of relatedness. There is no difference in the total activity of these units when integration (strong ventriloquism) happens and when segregation (no ventriloquism) happens. This could possibly explain why Rohe and Noppeney did not find very strong correlation of $\hat{S}_{A,c=2}$ and $\hat{S}_{V,c=2}$ with activation of IP1-2. Finally, the reconstruction layer contains unisensory auditory and visual units that generate activity profiles, which encode the final estimates of auditory and visual stimulus locations, respectively. The anterior region of intraparietal cortex (IP3-4) shows activation patterns that are also highly correlated with the final estimates of for auditory and visual locations $\hat{S}_A$ and $\hat{S}_V$.

Multisensory recalibration of auditory space, as measured by the strength of the ventriloquism aftereffect, correlates with elevated activation of the planum temporale (PT) on the hemisphere contralateral to the direction of the sound shift (Zierul et al., 2017). In our model, the ventriloquism aftereffect results from the up-regulation (down-regulation) of input weights onto the input units that produced less (more) activity then what their reconstruction unit counterparts predicted. This adjustment of weights produces subsequent elevated activation of auditory input units whose direction of tunining matches the direction of the previously presented audiovisual disparity (adaptation stimulus); such a mechanism could explain the findings by Zierul et al. (2017). Due to the sigmoidal tuning properties of the auditory input units and the weight update rule that we have proposed, our model predicts that the auditory input units whose RF centers ($x_i$) are near the mid-line should be adjusted (up or down-regulated) the most.

## 4.2 Relation to other neural network models

In the introduction, we briefly mentioned a number of network models that have inspired the one detailed in this paper (Ma & Rahmati, 2013; Spratling, 2016; Zhang et al., 2016; Magosso, Cuppini, & Ursino, 2017). Here, we will discuss in more detail the similarities and differences between some of these models and ours. The Predictive Coding, Biased Competition and Divisive Input Modulation (PC/BC-DIM) model proposed by Spratling (2016) is a main inspiration for the architecture and feedback mechanism of our neural network model. The Spratling network also consists of 3 layers: an input layer, an intermediate layer and a reconstruction layer, with each layer playing a similar role to the layers of this network. The intermediate layer represents the underlying causes of the inputs to the network (the generative model(s)), and the reconstruction layer represents the predicted input given these generative models. In our model the intermediate layer consists of 3 populations; the auditory and visual unisensory populations represent the generative model that the auditory and visual inputs came from separate causes, while the multisensory population represents the generative model that auditory and visual inputs came from the same cause. Our reconstruction layer also represents the predicted activities of these generative models; because our model creates predicted inputs for both input populations, auditory and visual, there are two layers in our reconstruction layer. A big difference between our model and Spratling's (2016) is the role of feedback of the reconstructed inputs. Our model uses the reconstructed activity as a template for unsupervised recalibration of input units, while Spratling uses the reconstructed activity as a prediction signal to calculate prediction error and to propogate the error signal forward in a recursive manner. Our model only takes a single pass through the network to generate the final estimate, whereas Spratling takes multiple (tens) of cycles to converge to a final estimate. The types of stimuli our model processed for the simulations we carried out are relatively quick in duration (tens of milliseconds), static and local. Therefore we assume a single pass through the network is sufficient to generate an estimate.

Another neural network model that greatly influences ours is Zhang et al, 2016. In their model, separate populations of units known as "congruent" and "opposite" neurons are maximally activated by aligned (0 degree disparity) and misaligned (180 degrees disparity) inputs for the heading direction. These populations together represent the probability that both types of inputs (visual and vestibular) are either consistent with a congruent heading direction or that they are consistent with opposite heading directions. Our model is similar in that the intermediate layer makes use of two types of units (multisensory and unisensory) that represent two possible relationships between stimuli: either they share the same cause and come from the same location, or they are independent and therefore more likely to come from separate locations. Our unisensory units are quite different from their "incongruent" neurons, since they do not fire maximally when stimuli are far apart. Rather, they simply represent the likelihood of a unisensory stimulus' position, and the firing rates of these units do not vary with audiovisual disparities. Instead the relatedness of two stimuli is reflected in the activity of the multisensory units relative to the unisensory units.

Magosso, Cuppini, & Ursino's (2017) network model contains dynamic weights learned through experience and, therefore, models the development of a proposed circuitry for causal inference. Our model, however, does not explain the development of such circuits but rather proposes a simple architecture capable of approximating causal inference. Their model, like many others, also uses topographically organized Gaussian tuning curves for their auditory units. However, most auditory cortical neurons, as well as many unisensory SC auditory neurons, possess a rate code of (sigmoidal tuning for) space rather than a place code; this is an attribute that our model captures.

### 4.3 Assumptions, limitations and future work

One major limitation and simplification of our network is that it does not model temporal dynamics at the level of neural activity; we only model time on a trial-by-trial basis (with a resolution of 1 second) for the recalibration/adaptation of input weights. Rather, our units represent idealized neurons that simply take weighted sums of activity to imitate the summation of postsynaptic potentials followed by a non-linear (exponential) transformation to imitate spiking behavior and neural-transmission. By implementing models of spiking neurons, or recursive units, we would be able to address questions of temporal significance. For example, timing of inputs is important for estimation by the causal inference model, as the ventriloquism effect has been demonstrated to be dependent on the timing of stimuli (McGovern et al. 2016). Furthermore, the timing of spikes is hypothesized to play an important role in multisensory integration (Bieler et al, 2017). Spike timing dependent plasticity may also offer a more biologically plausible way of modeling the adaptation of our input weights through error minimization (Bengio et al, 2015).

We have chosen not to model non-uniform priors over location to keep our focus on the most basic version of Causal Inference, and to avoid over-complicating the model. However, since intrinsic biases in spatial localization have been demonstrated to exist in behaviour and have been modeled as Gaussian priors peaked over certain locations (Odegaard, Wozny & Shams, 2015; Körding et al. 2007), further steps can be taken to account for non-uniform position priors that would lead to an overall stronger model. This can be implemented by using the same, or similar, approach as Spratling (2016) to model non-uniform priors over position: the weight distributions between input and intermediate layer are scaled by a Gaussian function that reflects the prior distribution over space. Indeed, this method was demonstrated to be successful in modeling a prior over space.

We have delegated the role of determining the prior relatedness of multisensory stimuli to a hypothetical external network, which in turn feeds activity into our multisensory pooling units. Although we have gone beyond treating this initial activity, $\mu$, as a free parameter and actually determined a relationship between $p_{common}$ and the $\mu$ needed to model such a $p_{common}$ (see *Eq. 17*), a proper network model of how inputs could be transformed to produce the desired output should be proposed in the future. One possible neural substrate to gain insight from is the Superior Temporal Sulcus, a structure in the ventral processing stream, which is thought to encode the semantic congruency or contextual relatedness of multisensory stimuli (Werner & Noppeney, 2010). Further evidence shows that the ventral pathway (including the STS) is recruited earlier than the dorsal pathway during landmark cueing tasks, in which saccades are to be made to a spatial target (Lambert & Wootton, 2017). Therefore, dorsal pathway processing might generate prior expectations that are useful for spatial localization, as we have proposed here. A network model of this processing, in combination with the model proposed here, might link multisensory processing of ventral and dorsal processing streams into a more unified model of perceptual binding.

## 4.4 Conclusion

In conclusion, we have presented a neurocomputational model of multisensory causal inference and recalibration that proposes unique roles for unisensory and multisensory neurons along the dorsal processing hierarchy. We use inputs modeled after single-unit responses from auditory and visual cortices in a neural architecture inspired by the connection patterns between lower-level unisensory, and higher-level multisensory cortices. We demonstrated that the network model approximates the mean behavior generated by the causal inference model (Körding et al., 2007) and lastly, that it reproduces the phenomenon of auditory recalibration (ventriloquism aftereffect), and its time course of accumulation and decay.

Tables

**Table 1.** Parameter settings used for simulations.

| Parameter | Value(s) used for simulations |
|---|---|
| $A$<br>*( auditory weight scaling factor)* | 2 |
| $V$<br>*( visual weight scaling factor)* | 5 ,<br>4.335 *( for figure 10)* |
| $V_m$<br>*( visual weight scaling factor for multisensory units)* | 2 |
| $n_L$<br>(number of leftward auditory units) | 301 |
| $n_R$<br>(number of rightward auditory units) | 301 |
| $n_V$<br>(number of visual units) | 301 |
| $L$<br>*(range of locations covered by units)* | 301 |
| $\alpha_i$<br>*(adaptation weights)* | 1,<br>$\alpha_i \in \mathbb{R}$ *(recalibration simulations)* |
| $m$<br>*(rise parameter for auditory input)* | 20 |
| $\sigma$<br>*(width of visual input tuning curve)* | 20<br>80 *(for figure 10)* |

Unless stated (see parentheses), the above values were the default values used in simulations. See discussion section 4.3 for a justification of weight-scaling parameters.

## Acknowledgments

This research was supported by funding from the German Research Foundation (DFG) TRR 169 awarded to Prof. Dr. Stefan Wermter and Prof. Dr. Brigitte Röder.